\documentclass[letterpaper, 10 pt, conference]{ieeeconf}
\usepackage{cite}
\usepackage{amsmath,amssymb,amsfonts} 
\usepackage{graphicx}
\usepackage{textcomp} 
\usepackage{multirow}
\usepackage{gensymb}
\usepackage{xcolor} 
\usepackage[ruled,vlined]{algorithm2e}
\usepackage{gensymb}
\usepackage{textcomp} 
\usepackage{xcolor} 
\usepackage{color}
\usepackage{multirow}
\usepackage{graphicx}
\usepackage{subfigure}
\usepackage[mathscr]{eucal}
\usepackage{makecell}

\IEEEoverridecommandlockouts                              

\makeatletter

\makeatletter
\renewcommand{\maketag@@@}[1]{\hbox{\m@th\normalsize\normalfont#1}}%

\setbox0\hbox{$\xdef\scriptratio{\strip@pt\dimexpr
    \numexpr(\sf@size*65536)/\f@size sp}$}

\newcommand{\scriptveryshortarrow}[1][3pt]{{%
    \vcenter{\hbox{\rule[\scriptratio\dimexpr-.2pt\relax]
               {\scriptratio\dimexpr#1\relax}{\scriptratio\dimexpr.4pt\relax}}}%
   \mkern-4mu\hbox{\let\f@size\sf@size\usefont{U}{lasy}{m}{n}\symbol{41}}}}

\makeatother

\title{\LARGE \bf
Automatic Surround Camera Calibration Method in Road Scene for Self-driving Car
}

\author{Jixiang Li, Jiahao Pi, Guohang Yan$^{\dagger}$ and Yikang Li 
\thanks{$^{\dagger}$ Corresponding author.}
\thanks{Jixiang Li, Jiahao Pi, Guohang Yan and Yikang Li are with Autonomous Driving Group, Shanghai AI Laboratory, China. {\tt\small \{yanguohang, hefeiyu, liyikang\}@pjlab.org.cn}}
}
    
\begin{document}
 
\maketitle

\begin{abstract}
With the development of autonomous driving technology, sensor calibration has become a key technology to achieve accurate perception fusion and localization. Accurate calibration of the sensors ensures that each sensor can function properly and accurate information aggregation can be achieved. Among them, camera calibration based on surround view has received extensive attention. In autonomous driving applications, the calibration accuracy of the camera can directly affect the accuracy of perception and depth estimation.
For online calibration of surround-view cameras, traditional feature extraction-based methods will suffer from strong distortion when the initial extrinsic parameters error is large, making these methods less robust and inaccurate.
More existing methods use the sparse direct method to calibrate multi-cameras, which can ensure both accuracy and real-time performance and is theoretically achievable. However, this method requires a better initial value, and the initial estimate with a large error is often stuck in a local optimum.
To this end, we introduce a robust automatic multi-cameras (pinhole or fisheye cameras) calibration and refinement method in the road scene. We utilize the coarse-to-fine random-search strategy, and it can solve large disturbances of initial extrinsic parameters, which can make up for falling into optimal local value in nonlinear optimization methods. In the end, quantitative and qualitative experiments are conducted in actual and simulated environments, and the result shows the proposed method can achieve accuracy and robustness performance. The open-source code is available at https://github.com/OpenCalib/SurroundCameraCalib.
\end{abstract}

\section{INTRODUCTION}
The development of autonomous driving technology, especially the recent BEV (Bird's Eye View) surround view perception algorithm, has become the key to upgrading intelligent driving technology, and more and more surround view camera solutions are being used. 
In the surround-view fisheye, cameras are usually used for close-range perception in autonomous driving. Four fisheye cameras on the four sides of the vehicle are sufficient to cover a 360° range around the car and capture the entire short-range area. Some application scenarios include parking-slot detection \cite{zhang2018vision}, autonomous parking \cite{lin2012vision} and pedestrians detection \cite{gressmann2011surround}, etc.
\begin{figure}[htbp]
\centering
    \includegraphics[width=0.5\textwidth,height=0.25\textwidth]{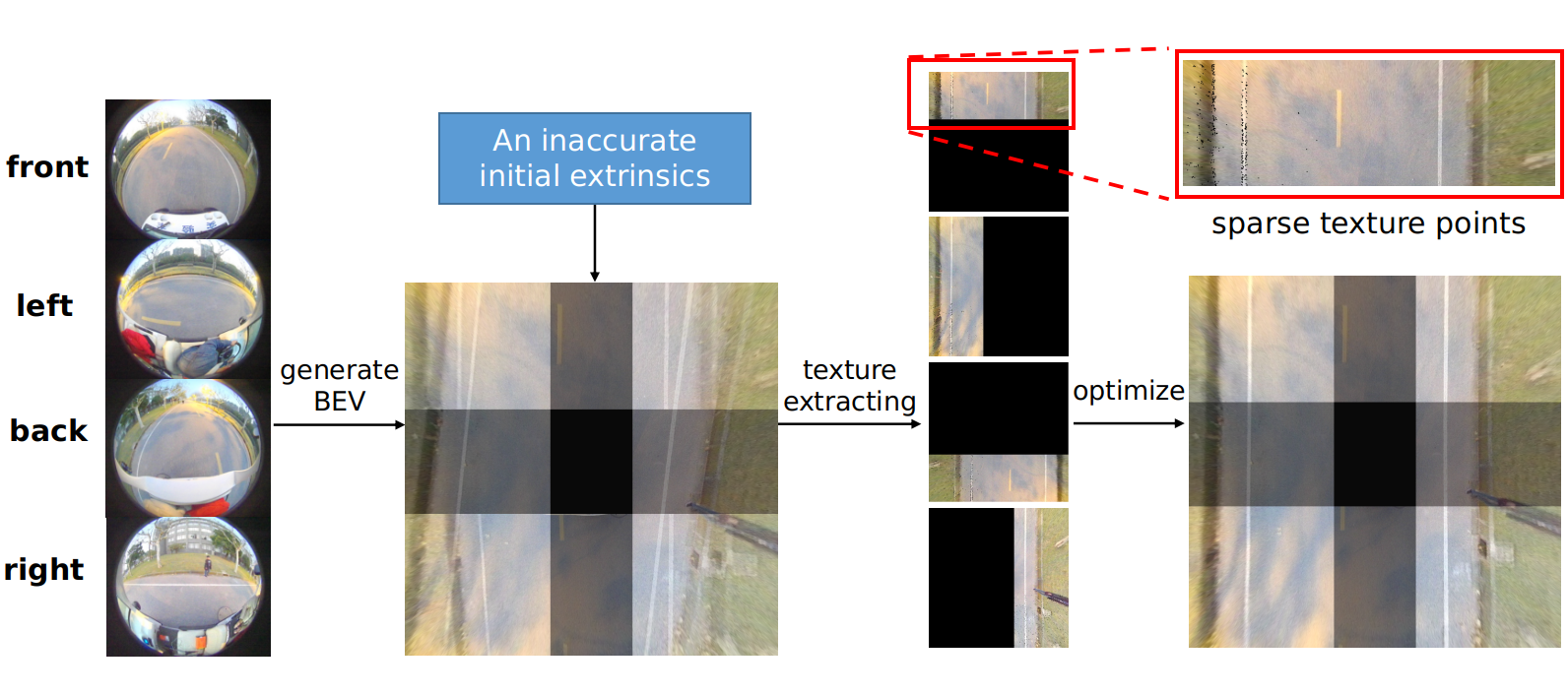}
    \caption{Overview of the presented method. First, we convert the images to BEV according to the initial extrinsic parameters. Then calculate the common area between adjacent cameras, extract the texture pixel points under BEV, project the texture back to the image of another camera and the computer photometric error, and obtain the final extrinsic parameters that minimize the photometric loss in the coarse-to-fine random search stage result.}
    \label{fig:overview}
\end{figure}
For surround-view camera systems, the accuracy in intrinsic and extrinsic parameters plays an essential role in these driving assistance applications. When the self-driving or assisted driving vehicle is off the production line, the production line calibration will be carried out, and the intrinsic parameters of the cameras and the extrinsic parameters of the surround-view camera will usually be calibrated. However, during the driving process of the vehicle, due to vibration and other reasons, the position and angle of the sensor will deviate from the original pose, so it is necessary to calibrate the sensor at regular intervals. The paper presents a novel surround-view camera extrinsic method in road scenes, which is a robust, accurate, online calibration method for multi-cameras. The method can be easily embedded into autonomous cars in practical application scenarios. Our method do not rely on the strong assumption, like in lane-line based methods \cite{collado2006self,nedevschi2007online}, which need two parallel line on ground captured by multi-cameras.The contributions of this work is listed as follows: 
\begin{enumerate}
\item We propose a fully automatic and targetless method for surround-view cameras extrinsic calibration based on photometric errors in overlapping regions under bird's eye view in road scenes.
\item We use a coarse-to-fine random search strategy, which can adapt to large initial extrinsic errors. At the same time, avoid the problem of getting stuck in the optimal local value in the nonlinear optimization method.
\item The proposed method shows promising performance on our simulated and real-world data sets; meanwhile, based on the analysis, we develop a practical calibration software and open source it on GitHub to benefit the community.

\end{enumerate}

\section{RELATED WORK}

\subsection{Lane-based Methods}
The lane-line based methods rely on a strong assumption, that is two parallel lane-lines on the ground can be captured by the cameras. In \cite{collado2006self}, Collado et al.  used the Sobel operator and the Hough transform to extract the calibration pattern from two parallel ground lanes, and then with the pattern, they estimated the extrinsics of the stereo-vision cameras. S. Nedevschi et al. in \cite{nedevschi2007online} first estimates the vanishing point based on two lanes parallel to each other on the flat ground, and then with the estimated vanishing point, the pose of the multi-camera system relative to the world coordinate system is solved.  Hold et al. in \cite{hold2009novel} proposed a method of online extrinsics calibration, which also use ground lanes was proposed. Firstly, they detected the lane-line and obtained a series of feature points by sampling the lane with the scanning line. Then, fast Fourier transform was adopted to measure the distance of lane points, and finally, they made use of lane points to solve the cameras’ extrinsics. In \cite{zhao2014automatic} Zhao et al. proposed to utilize multiple vanishing points of lane markings for calibrating cameras’ orientations. These aforementioned lane-line based methods are all not applicable to the SVS case. In \cite{choi2018automatic}, Choi et al. proposed a method to calibrate SVS by aligning marking across images of adjacent cameras, but it still make a assumption of two parallel lane-line.

\subsection{Odometer-based Methods}
The odometry methods take calibration questions into optimization of visual odometry or a complete SLAM system so as to the extrinsic can be optimized by the mean time. In \cite{schneider2013odometry}, Schneider et al. proposed a pipeline of calibrating extrinsics of cameras which resorted to the localization of a visual odometry, but it has a drawback that it is need about 500 frames for the system to converge. In \cite{heng2015leveraging, heng2013camodocal}, Heng et al. introduced a infrastructure-based pipeline to calibrate cameras. If SVS use this scheme, the car equiped multi-cameras needs to travel around the calibration area for a short while to establish the map. However, since the SLAM system need a significant time and computational resources to converge. these methods are not applicable for the portability of industrial SVS's online calibration.

\subsection{Photometric-based Methods}
In the direct image alignment method, the local intensity is utilized to determine the step of the optimization first in \cite{forster2014svo}, All images pixels or most of its pixels can be participated in optimization, which, so demonstrate better robustness in scenes with sparse textures. Nowadays more and more researchers are willing to adopt direct methods rather than feature-point based ones to recover camera poses from images, especially in the field of SLAM\cite{irani2000direct,newcombe2011dtam,engel2014lsd,engel2017direct,mur2017orb}. In the field of calibration of SVS, direct method are also be utilized in the recent times, such as Liu et al. first proposed the self-calibration schemes utilized direct methods in\cite{liu2019online}. And In \cite{zhang2020oecs}, Zhang et al.  designed a novel model, the bi-camera model, to construct the least-square errors\cite{lourakis2010sparse} on the imaging planes of two adjacent cameras, and then optimize camera poses by the LM (Levenberg-Marquardt) algorithm\cite{more2006levenberg}. In \cite{zhang2021roecs} they further improved their work in \cite{zhang2020oecs} by utilizing multiple frames selected and stored in a local window rather than a single frame to build the overall error, so as to improve the system’s robustness. Since the above three studies \cite{liu2019online,zhang2020oecs,zhang2021roecs} focused on the “online correction” rather than the “calibration”, a rough initial extrinsics needed to be offered to them as the input. The same goes for our schemes. But aforementioned direct based calibration schemes are also based on non-linear optimization methods, which can make algorithm fall into local optimal value, so these methods can not solve relatively large disturbance. 

\section{METHODOLOGY}
This section introduces the details of our approach, including texture points extraction, optimization loss and The coarse-to-fine solution.
Fig.~\ref{fig:overview} shows the overview of the presented method.
\subsection{Texture points extraction and optimization loss}
This section describes how to generate BEV images, extract textures, project back texture pixels, and optimize loss.

\subsubsection{Projection model}
 Given the ground coordinate system $O_G$ (BEV's coordinates) and a surround-view system consisting of four cameras $C_1$, $C_2$, $C_3$ and $C_4$, the poses of cameras in $O_G$ are denoted by $T_{C1G}$, $T_{C2G}$, $T_{C3G}$ and $T_{C4G}$, respectively. For a point $P_G$ = [$X_G$,$Y_G$,$Z_G$]$^T$ in $O_G$ , its corresponding pixel coordinate, take our pipeline, for example, $p_{Cj}$ = [$u_j$,$v_j$,1] (j=1,2,3,4) in the camera coordinate system of $C_j$, the relationship is given by,

\begin{equation}
\begin{aligned}  
p_{Cj}=\frac{1}{Z_{Cj}}K_{Cj}T_{C_jG}P_G,i=1,2,3,4
\end{aligned}
\end{equation}

where $Z_{Cj}$ is the depth of $P_G$ in camera $C_j$ ’s coordinate system, and $K_{Cj}$ is the 3×3 intrinsic matrix of camera $C_j$, which can be estimated by Zhang's work \cite{zhang1999flexible}.  It's worth to mention that when we use fisheye camera, the intrinsic is calibrated by OCamCalib \cite{scaramuzza2006toolbox}.
And $P_G$ is BEV points which generated by BEV camera whose camera parameters can be set artificially, we use $K_G$ as its camera matrix, so projection formula is given by,
 \begin{equation}
\begin{aligned}  
P_G=K_G^{-1}p_G 
\end{aligned}
\end{equation}
where $p_G$ = [$u_g$,$v_g$,1] is BEV image's pixels, which can be generated by camera $C_j$'s adjacent camera like $C_i$. By combining Equation (1) and Equation (2), we can get,
 \begin{equation}
\begin{aligned}  
p_{Cj}=\frac{1}{Z_{Cj}}K_{Cj}T_{C_jG}K_G^{-1}p_G,i=1,2,3,4
\end{aligned}
\end{equation}
Due to we do not know the depth of points in the surround cameras' view, so we can not directly project the surround camera image's pixels into 3d points, so we can not get the 3d points and their corresponding 2d pixels on the ground, but in BEV we suppose the depth of ground points are same which are equal to BEV camera's height to the ground. So we can project back every pixel's coordinate back to camera, for example, the aforementioned camera $C_i$ in our pipeline (just like Equation (3) describes, but replace index j to index i). We get the corresponding pixel coordinate in camera $C_i$. So we can use this mapping relation to get BEV image of camera $C_i$, these can be implemented by OpenCV\cite{opencv_library} library.

\subsubsection{Texture points extraction}
It's worth mentioning that the texture points we extract are only in the common-view of two adjacent cameras, which causes the texture points beyond the common-view are not contributed to our optimization. As we know, if the stuff on the ground can be seen by two adjacent cameras in their BEV images, the corresponding pixels will be appear in the same image area, if the poses of these two camera are very precise, these textures pixels will be at the same location (share the same coordinate). Based on this multi-view geometry prior knowledge and initial calibration parameters, we can project one camera's BEV texture pixels back to the other's camera's image and computer photometric loss, and we just optimize this loss to make two cameras BEV textures to be overlapped. 

\begin{figure}[htbp]
\centering
    \includegraphics[width=0.4\textwidth,height=0.36\textwidth]
{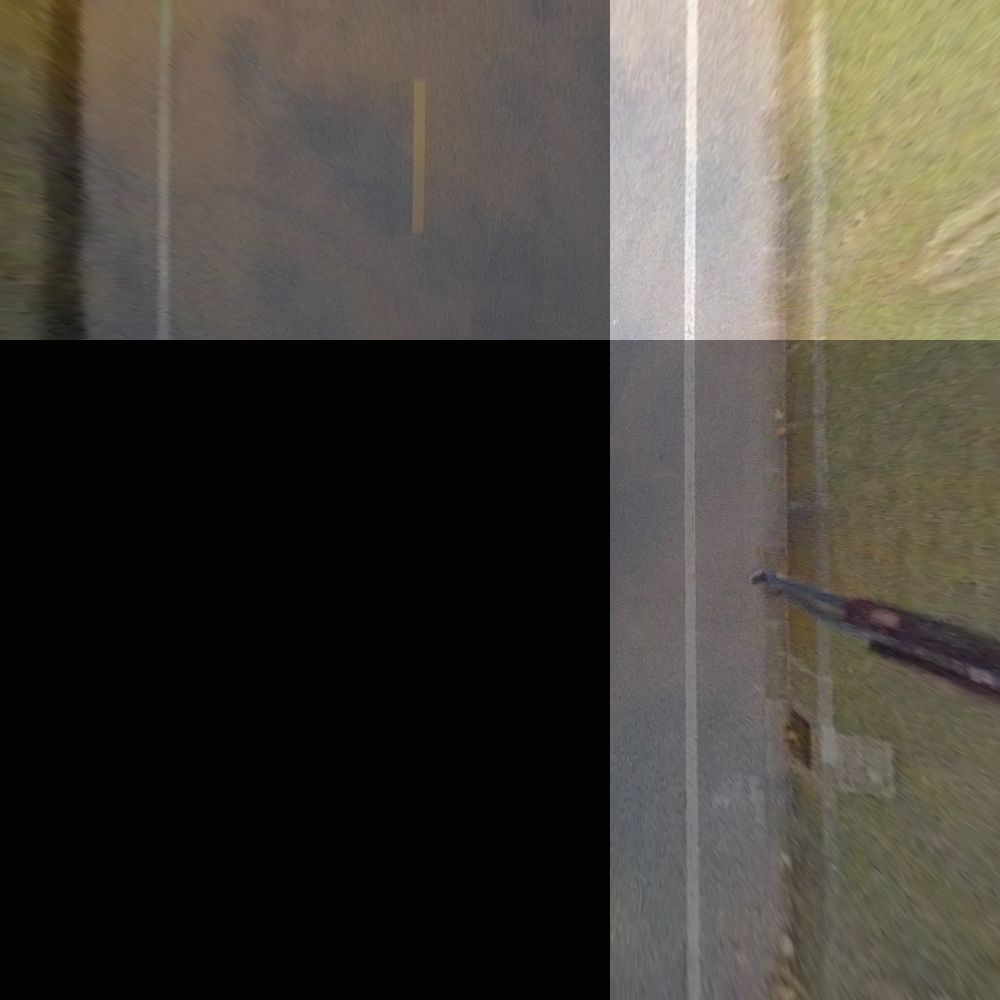}
    \caption {Combination of BEV images of two adjacent cameras after optimization. After the optimization of the right camera, the white lane-line and curb gap could be overlapped well.}
    \label{fig:camera_projection}
\end{figure}

At first, we project two camera's images to BEV images with their initial poses as shown in Fig.~\ref{fig:camera_projection}. 
Then, we compute and get the mask of common-view, which can be detected contour by OpenCV\cite{opencv_library} library. 
In this region, we perform texture point extraction.
There we introduce the pixel's photometric gradient, which can be taken as a 2D vector.

\begin{equation}
\begin{aligned}  
\Delta=(gray[x][y]-gray[x][y-\delta], 
\\ gray[x][y]-gray[x-\delta][y])
\end{aligned}
\end{equation}
if gradient's L2 norm satisfy:
\begin{equation}
\begin{aligned}  
\Vert \Delta \Vert_2 < \theta
\end{aligned}
\end{equation}
we consider this pixel point as a texture pixel point. In our algorithm, we assign $\delta$ value with 2, $\theta$ value with 15. It's worth mentioning that if adjacent cameras have different exposure conditions, we should take measures to solve this. In \cite{liu2019online,zhang2020oecs,zhang2021roecs}, these methods define a factor as the ratio of exposure time of $C_i$ and $C_j$:
\begin{equation}
\begin{aligned}  
\gamma_{ij}=\frac{t_i}{t_j}
\end{aligned}
\end{equation}
where $t_i$ is $C_i$'s exposure time and $t_j$ is that of $C_j$'s. The exposure time can be obtained with the photometric calibration introduced in\cite{engel2016photometrically}. In above three methods, they offer a simple scheme for approximation, in which the factor can be fitted as, 
\begin{equation}
\begin{aligned}  
\gamma_{ij}=\frac{\sum{p_G\in O_{ij}I_{GC_i}(p_G)}}{\sum{p_G\in O_{ij}I_{GC_j}(p_G)}}
\end{aligned}
\end{equation}
Having the common-view region in BEV view, we compute the $\Delta$ in this region and compare it with the threshold. Finally, we get the texture pixel points of camera's BEV image.

\subsubsection{Compute photometric loss}
We get the BEV texture pixel points of camera $C_i$. Then need to project these texture pixel points to camera $C_j$'s image and get the corresponding pixel points, and then compute the photometric loss between these two camera views, that's camera $C_i$'s BEV image view and camera $C_j$ view, which is given by,
\begin{equation}
\begin{aligned}  
\varepsilon_{p_G}=\Vert I_{GC_i}(p_G)-I_{C_j}(p_j) \Vert_2
\end{aligned}
\end{equation}
where $I_{GC_i}$ is camera $C_i$'s BEV image generated by pose $T_{GC_i}$, $I_{C_j}$ is camera $C_j$'s image,  $p_G$ is texture pixel coordinate in $C_i$'s BEV image($I_{GC_i}$), and $p_j$ is $p_G$'s corresponding pixel coordinate in $C_j$'s image. We expanding Equation(8) into equation with pose $T_{C_jG}$ by combining Equation(3),
\begin{equation}
\begin{aligned}  
\varepsilon_{p_G}=\Vert I_{GC_i}(p_G)-I_{C_j}(\frac{1}{Z_{Cj}}K_{Cj}T_{C_jG}K_G^{-1}p_G) \Vert_2
\end{aligned}
\end{equation}
where $K_{C_j}$ is camera matrix of camera $C_j$, $K_G$ is camera matrix of camera BEV. $Z_{Cj}$ is depth of 3D point in camera $C_j$'s coordinate system, which is projected from BEV.
\subsection{The coarse-to-fine solution}
The best calibration results are those that aims at generating seamless BEV surround-view image, 
It should be noted that the front view camera can calibrate the extrinsic parameters from the camera to the car body through the vanishing point and the horizontal line \cite{yan2023sensorx2car}.
It should be noted that the front view camera can calibrate the extrinsic parameters from the camera to the car body through the vanishing point and the horizontal line \cite{yan2023sensorx2car}.
Then fix the front camera and optimize all remaining cameras mounted on the car by recursively optimizing photometric loss in every common-view region of two adjacent cameras.


Since image alignment based on photometric loss optimization is non-convex and can hardly be optimized by convex optimization techniques, If using the optimized solution, you need to provide a reasonable and robust initial extrinsic estimation. In this way, the large initial extrinsic error cannot be corrected. To solve this problem, we utilize the coarse-to-fine random search strategy; the photometric loss is optimized by randomly searching the parameter space around the current optimal parameter. 

In the grading random-search pipelines, in each phase, we set the search scope of the parameters, which are $\delta_{roll}$, $\delta_{pitch}$, $\delta_{yaw}$, dx, dy, dz. For $\delta_{roll}$, $\delta_{pitch}$, $\delta_{yaw}$, we define disturbance optimization search scope as,
\begin{equation}
\begin{aligned}  
A < \delta_{roll},\delta_{pitch},\delta_{yaw} < B 
\end{aligned}
\end{equation}
for dx, dy, dz, we define disturbance optimization search scope as,
\begin{equation}
\begin{aligned}  
\alpha< dx,dy,dz < \beta
\end{aligned}
\end{equation}
where $A$, $B$ is rotation search range, and $\alpha$,$\beta$ is translation search range. In each round of the first random-search phase, we can get a temporary pose $T_{C_jG}^{1st}(k)$ , which can also generate a BEV image.
\begin{equation}
\begin{aligned}  
T_{C_jG}^{1st}(k)=\partial T_{C_jG}T_{C_jG}^{ini}
\end{aligned}
\end{equation}
where k represent the kth round of random-search in the first phase. And at each round we use $T_{C_jG}^{1st}(k)$ to compute photometric loss, 
\begin{footnotesize}
\begin{equation}
\begin{aligned}  
&\sum_{p_G\in N_{ij}}\varepsilon_{p_G}^2(k)=\\
&\sum_{p_G\in N_{ij}}\Vert I_{GC_i}(p_G)-I_{C_j}(\frac{1}{Z_{Cj}}K_{Cj}{T_{C_jG}^{1st}}(k)^{-1}p_G) \Vert_2
\end{aligned}
\end{equation}
\end{footnotesize}
where $N_{ij}$ is texture pixels in BEV image's common-view of two adjacent cameras, such as camera $C_i$ and $C_j$, and if the photometric loss sum is less than the BEV image generated by the optimal pose $T_{C_jG}^{optimal}$, 
\begin{footnotesize}
\begin{equation}
\begin{aligned}  
&\sum_{p_G\in N_{ij}}\varepsilon_{p_G}^2(k)<\\
&\sum_{p_G\in N_{ij}}\Vert I_{GC_i}(p_G)-I_{C_j}(\frac{1}{Z_{Cj}}K_{Cj}{T_{C_jG}^{optimal}}^{-1}p_G) \Vert_2
\end{aligned}
\end{equation}
\end{footnotesize}
we update the $T_{C_jG}^{optimal}$ by $T_{C_jG}^{1st}(k)$,
\begin{equation}
\begin{aligned}  
T_{C_jG}^{optimal}=T_{C_jG}^{1st}(k)
\end{aligned}
\end{equation}
and in the second and third random-search phase, we multiply $\partial T_{C_jG}$ by the $T_{C_jG}^{optimal}$, 
\begin{equation}
\begin{aligned}  
T_{C_jG}^{2nd/3rd}(k)=\partial T_{C_jG}T_{C_jG}^{optimal}
\end{aligned}
\end{equation}
same as before, if this round's photometric loss sum is less than BEV image generated by the optimal pose $T_{C_jG}^{optimal}$, we update it by $T_{C_jG}^{2nd/3rd}(k)$.

It's worth to mention that we introduce this random-search strategy, which is a bit similar to the gradient descent methods in non-linear optimization, but the strategy can avoid falling into local optimal value. That is when in the second or third phase of random search, as long as we find a relative parameter that can generate BEV image with less photometric loss in the common-view region compared to the temporary optimal pose $T_{C_jG}^{optimal}$, we immediately update it, at the same time we continue to random-search within the scope set by the algorithm around the updated pose($T_{C_jG}^{optimal}$)  in this phase, even if we are at the optimal value, cause the search scope of our algorithm is not given by gradient, and the scope is definitely large than the step around optimal value computed by gradient, so we can deviate optimal value.

But we do not take this measure in the first random-search phase cause in the first phase, the correction(for example, $\partial T_{C_jG}$) is relatively large than the other two phases; taking this measure will risk making algorithm even get the pose which is far from the ground-truth, so in the first phase we randomly search the coarse pose in the fixed scope.

\section{EXPERIMENTS}
The experiment in this paper consists of two parts: a realistic experiment on our driverless vehicle test platform and a simulated experiment based on the Carla engine \cite{Dosovitskiy17}.
\vspace{-1mm}
\subsection{Experiment Settings}
\begin{figure}
\centering
\includegraphics[width=0.5\textwidth,height=0.3\textwidth]{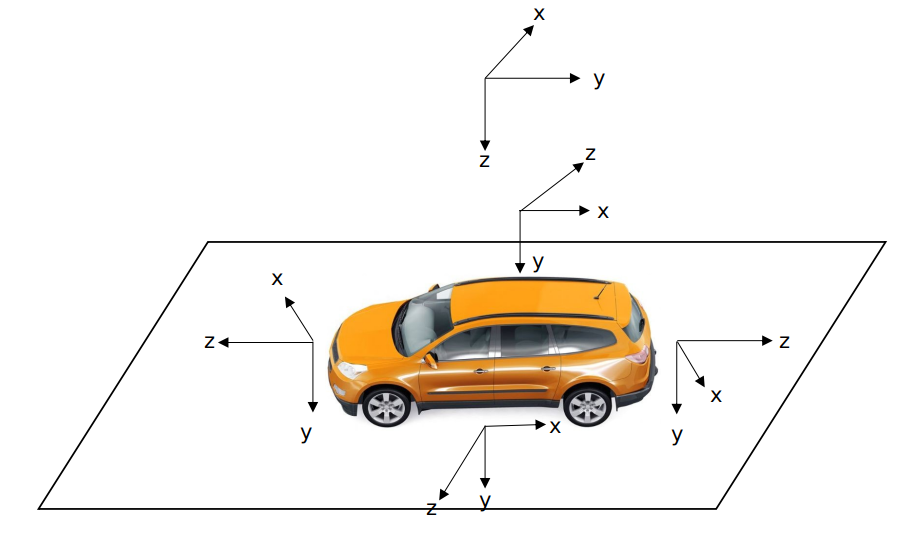}
    \caption{Surround cameras' poses and BEV camera's pose}
 \label{fig:car_setting}
\end{figure}
\textbf{Machine configuration:} We execute our algorithm on Ubuntu 20.04.5 LTS with the processor(11th Gen Intel® Core™ i7-11700 @ 2.50GHz × 16). The installation position of the surround view camera is shown in Fig.~\ref{fig:car_setting}

\begin{figure}[htbp]
\centering
\includegraphics[width=0.5\textwidth,height=0.15\textwidth]{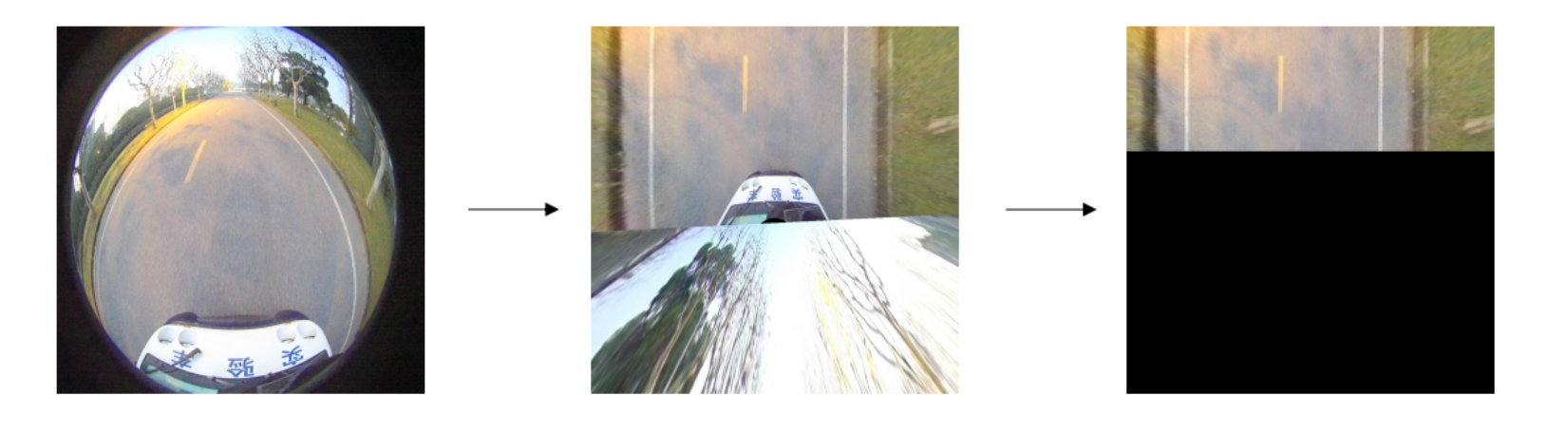}
    \caption{pipeline of generating BEV view and cutting of the surrounding environment.}
 \label{fig:bev_tail}
\end{figure}

Our algorithm use two datasets: first is Carla simulated data(4 cameras mounted at each side of the car, Fov=125\degree, resolution: 1500*1500) on normal street with some ground textures, such as tiles, lanes and so on. second is  fisheye camera data collected from car in real world (Fov=195\degree, resolution:1280*1080). We filter the BEV map by ROI (region of interest) to remove the area occupied by the self-vehicle, as shown in Fig.~\ref{fig:bev_tail}.

\subsection{Qualitative Results}
\begin{figure}[htp]
\centering
    \includegraphics[width=0.5\textwidth,height=0.8\textwidth]{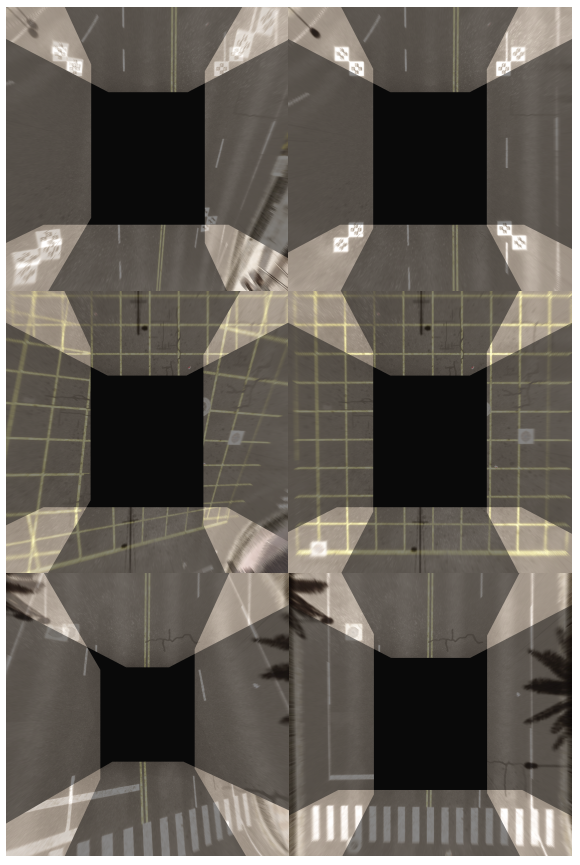}
    \caption{The result surround-view BEV images of the pinhole camera in Carla (FOV=125\degree), from top to bottom, is Scene 3,4,5. In the top case, we've added some calibration boards since the texture is barely visible in the small common areas in this scene.}
 \label{fig:carla_bev}
\end{figure}

\begin{figure}[htp]
\centering
    \includegraphics[width=0.5\textwidth,height=0.3\textwidth]{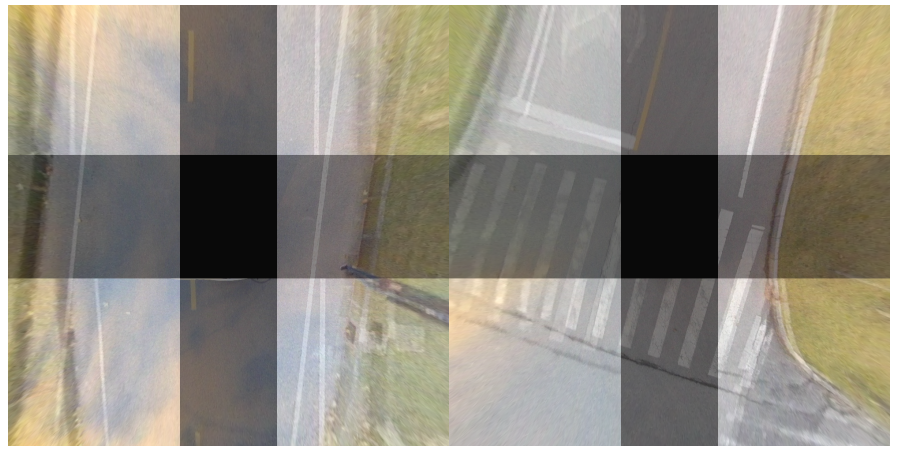}
    \caption{The initial surround-view BEV image before optimization. The left is Scene 1, and the right is Scene 2.}
 \label{fig:calib_before}
\end{figure}

\begin{figure}[htp]
\centering
    \includegraphics[width=0.5\textwidth,height=0.3\textwidth]{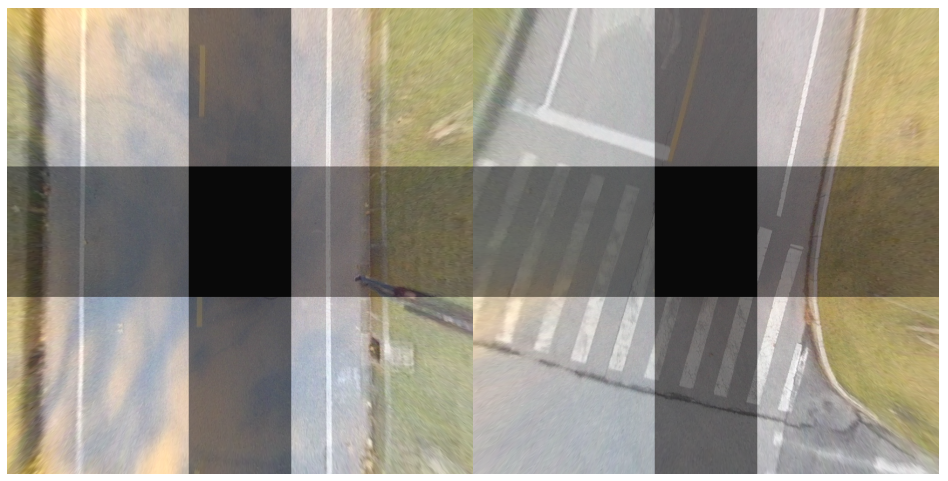}
    \caption{The final surround-view BEV image after optimization, the left is Scene 1, and the right is Scene 2.}
 \label{fig:calib_after}
\end{figure}

To better visualize the performance of our method, the four cameras are projected to BEV using the intrinsic and extrinsic parameters calibrated by our method. Results of the pinhole camera are shown in Fig.~\ref{fig:carla_bev}. Fig.~\ref{fig:calib_before} and Fig.~\ref{fig:calib_after} are results before and after calibration of the fisheye cameras. 

\subsection{Quantitative Results}
To the best of our knowledge, Liu et al.'s work\cite{liu2019online} Zhang et al.'s work\cite{zhang2020oecs,zhang2021roecs} are the existing three works that can solve the problem of online extrinsic correction for the surround-view systems with photometric loss. And \cite{zhang2021roecs} is the best in the three works, Thus we compare our work with it. 
The MAE is shown in Table \ref{table:result} and our calibration has a higher accuracy. After the initial error exceeds 0.3\degree, method \cite{zhang2021roecs} will fail to calibrate, but our method can still get better results. Our calibration algorithm can correct the initial error within 3\degree.

\begin{table}[tbp]
\vspace{-3mm}
\centering
\caption{MAE for Translation and Rotation of Our Method} 
\vspace{-3mm}
 \setlength{\tabcolsep}{2pt}
\renewcommand{\arraystretch}{1.2}
\begin{tabular}{|c|c|c|c|c|c|c|c|}
\hline
 Before   & camera & $t_x$(m) & $t_y$(m) & $t_z$(m) & roll(\degree) &pitch(\degree) &yaw(\degree) \\ \hline
GT &  \thead{left\\right\\rear}&\thead{ -1\\1\\0} & \thead{0\\0\\2} &\thead{4.1\\4.1\\4.1} &\thead{0\\0\\180}& \thead{90\\-90\\0}&\thead{-90\\-90\\90}      \\ \hline
 Calib   & camera & $\Delta t_x$(m) & $\Delta t_y$(m) & $\Delta t_z$(m) & $\Delta$\textit{roll}(\degree) &  $\Delta$\textit{pitch}(\degree) &  $\Delta$\textit{yaw}(\degree) \\ \hline
  initial error  &  \thead{left\\right\\rear}&\thead{ 0.095\\0.065\\-0.02} & \thead{0.025\\-0.075\\-0.076} &\thead{-0.086\\0.095\\0.096} &\thead{0.95\\-2.95\\-1.75}& \thead{1.25\\0.95\\2.95}&\thead{2.86\\2.8\\-1.8}     
 \\ \hline    
our&  \thead{left\\right\\rear}&\thead{ 0.015\\-0.008\\-0.019} & \thead{-0.008\\0.005\\-0.004} &\thead{-0.009\\ 0.020\\ -0.010} &\thead{-0.06\\0.08\\-0.19}& \thead{0.2\\-0.02\\-1.0}&\thead{-0.04\\0.14\\-0.38}     
\\ \hline
\end{tabular}
\label{table:result}
\end{table}

\section{CONCLUSIONS}
In this paper, we subvert the conventional methods, which include the feature-based or the direct-based by nonlinear optimization, to solve the online calibration of surround-view cameras, regardless of the fish-eye camera or pinhole camera. We propose a scheme that solves this problem by grading coarse-to-fine random-search methods. Not only can it make up the effect of distortion on feature-based methods, but also it can make up falling into optimal local values in nonlinear optimization when utilizing the direct-based methods. 
Most importantly, our method is still effective under large initial error.

In the future, we will try to optimize the time consumption of our method to improve the real-time performance of our algorithm, and at the same time improve the performance in the environment where the texture is not obvious.

\bibliographystyle{IEEEtran}
\bibliography{egbib}

\end{document}